\documentclass[preprint,12pt]{elsarticle}

\usepackage[T1]{fontenc}
\usepackage[utf8]{inputenc}
\usepackage{amssymb}
\usepackage{amsmath}
\usepackage{cancel} % For crossed-out math symbols
\usepackage{subfig}
\usepackage{pgf}  % For the PGF plots
\usepackage{multirow}
\usepackage{todonotes}
\usepackage{url}
\usepackage{hyphenat}
\usepackage{siunitx}
\usepackage{float}
\usepackage{algorithm}
\usepackage{algpseudocode}
\usepackage{mdframed}
\usepackage{enumitem}
\usepackage{fvextra}
\usepackage{inconsolata} % font to use for fixed width text
\usepackage{afterpage} % Make sure image and table are on the same page
\usepackage{makecell}
\usepackage{siunitx}
\usetikzlibrary{shapes.geometric, arrows.meta, plotmarks, positioning}

\newlength{\extralength}
\usepackage{booktabs} 
\usepackage{changepage}
\usepackage{multicol}

\usepackage[a4paper, left=2cm, right=2cm, top=1.5cm, bottom=3cm]{geometry}

\graphicspath{{./figs/}}

\makeatletter
\def\ps@pprintTitle{   \let\@oddhead\@empty
   \let\@evenhead\@empty
   \def\@oddfoot{\reset@font\hfil\thepage\hfil}
   \let\@evenfoot\@oddfoot
}
\makeatother

\graphicspath{{./Figures/}}

\begin{document}

\begin{frontmatter}

%% Title, authors and addresses

%% use the tnoteref command within \title for footnotes;
%% use the tnotetext command for theassociated footnote;
%% use the fnref command within \author or \affiliation for footnotes;
%% use the fntext command for theassociated footnote;
%% use the corref command within \author for corresponding author footnotes;
%% use the cortext command for theassociated footnote;
%% use the ead command for the email address,
%% and the form \ead[url] for the home page:
%% \title{Title\tnoteref{label1}}
%% \tnotetext[label1]{}
%% \author{Name\corref{cor1}\fnref{label2}}
%% \ead{email address}
%% \ead[url]{home page}
%% \fntext[label2]{}
%% \cortext[cor1]{}
%% \affiliation{organization={},
%%            addressline={},
%%            city={},
%%            postcode={},
%%            state={},
%%            country={}}
%% \fntext[label3]{}

\title{Time series forecasting based on optimized LLM for fault prediction in distribution power grid insulators}

\author[label1,label2,label3]{João P. Matos-Carvalho}
\author[label4]{Stefano Frizzo Stefenon}
\author[label5]{Valderi Reis Quietinho Leithardt}
\author[label4]{Kin-Choong Yow}

\affiliation[label1]{organization={LASIGE, Departamento de Informática, Faculdade de Ciências, Universidade de Lisboa, 1749--016 Lisboa, Portugal},%Department and Organization
%addressline={Campo Grande},
%city={Lisboa},
%postcode={1749-016},
%state={},
%country={Portugal}
}

\affiliation[label2]{organization={Center of Technology and Systems (UNINOVA-CTS) and Associated Lab of Intelligent Systems (LASI)},%Department and Organization
            %addressline={Caparica},
            city={Caparica},
            postcode={2829-516},
            %state={},
            country={Portugal}}

\affiliation[label3]{organization={COPELABS, Lus\'ofona University},%Department and Organization
            addressline={Campo Grande, 376},
            city={Lisboa},
            postcode={1749-024},
            %state={},
            country={Portugal}}

\affiliation[label4]{organization={Faculty of Engineering and Applied Sciences, University of Regina},%Department and Organization
            %addressline={Campo Grande, 376},
            city={Saskatchewan},
            postcode={S4S 0A2},
            %state={},
            country={Canada}}

\affiliation[label5]{organization={Instituto Universit\'ario de Lisboa (ISCTE-IUL), ISTAR},%Department and Organization
            %addressline={Campo Grande, 376},
            city={Lisboa},
            %postcode={S4S 0A2},
            %state={},
            country={Portugal}}

\begin{abstract}
Surface contamination on electrical grid insulators leads to an increase in leakage current until an electrical discharge occurs, which can result in a power system shutdown. To mitigate the possibility of disruptive faults resulting in a power outage, monitoring contamination and leakage current can help predict the progression of faults. Given this need, this paper proposes a hybrid deep learning (DL) model for predicting the increase in leakage current in high-voltage insulators. The hybrid structure considers a multi-criteria optimization using tree-structured Parzen estimation, an input stage filter for signal noise attenuation combined with a large language model (LLM) applied for time series forecasting. The proposed optimized LLM outperforms state-of-the-art DL models with a root-mean-square error equal to 2.24$\times10^{-4}$ for a short-term horizon and 1.21$\times10^{-3}$ for a medium-term horizon.  
\end{abstract}

%%Graphical abstract
%\begin{graphicalabstract}
%\includegraphics{grabs}
%\end{graphicalabstract}

%%Research highlights
%\begin{highlights}
%\item Research highlight 1
%\item Research highlight 2
%\end{highlights}

\begin{keyword}
%% keywords here, in the form: keyword \sep keyword

Deep learning \sep LLM; multi-criteria optimization \sep power grid insulator \sep time series forecasting

%% PACS codes here, in the form: \PACS code \sep code

%% MSC codes here, in the form: \MSC code \sep code
%% or \MSC[2008] code \sep code (2000 is the default)

\end{keyword}

\end{frontmatter}

%% \linenumbers

%% main text
\section{Introduction \label{INT}}

%\tableofcontents
\color{black}
Insulators in the conventional electrical power distribution and transmission system are installed outdoors and are susceptible to environmental conditions such as the presence of contamination, solar radiation, atmospheric discharges, and thermal variation~\citep{akbari2023experimental}. 
These environmental stresses can result in a power outage if preventive or predictive maintenance is not handled~\citep{castillo2021novel}. 

A way of identifying where the power grid needs more attention is to perform inspections on the electrical system, either through imaging or using specific equipment to assess the condition of the system~\citep{10559896}. 
Besides monitoring using specific equipment such as ultrasound, radio interference, ultraviolet, and infrared cameras, an analysis can be done in terms of power grid contamination. The contamination of insulators is a subject that has been extensively studied, because over time the contamination increases the surface conductivity of the insulators, resulting in flashover~\citep{qiao2021failure}.

Specific techniques for measuring the contamination of insulators, whether this is saline contamination or based on kaolin (artificial contamination) are explored. Measuring leakage current is one way of determining how the contamination is influencing the insulator's degradation~\citep{liu2020contamination}. The higher the surface conductivity of the insulator, the greater the chance of this component experiencing leakage current, which leads to a flashover voltage in the power system that can result in a complete shutdown of the grid~\citep{cui2020prediction}. 

Considering the advances made in predicting time series using machine learning (ML)-based models, especially involving deep learning (DL) \citep{yadav2024noa}, using them to predict the increase in leakage current could be an alternative for improving the monitoring performance of electrical systems. Based on this premise, this paper proposes a hybrid model for predicting leakage current in electric power distribution insulators. Given the vast nomenclature of the field, the acronyms of this paper are standardized according to Table \ref{acronyms}.

\begin{table}[]
%\footnotesize
%\small
\caption{Acronyms considered in this paper.}
\label{acronyms}
\centering
\begin{tabular}{p{2cm}p{9.5cm}p{6cm}}\hline
Acronym & Name\\ \hline
ANFIS &Adaptive neuro-fuzzy inference system\\
AR &Auto-regressive\\
A-LSTM & Attention-based long short-term memory \\
CF &Christiano Fitzgerald\\
CNN &Convolutional neural networks\\
DL & Deep learning\\
DeepNPTS &Deep non-parametric time series\\
DeepTCN &Deep temporal convolutional network\\
EI & Expected improvement \\
EMD &Empirical mode decomposition\\
EWT &Empirical wavelet transform\\
GMDH &Group method of data handling\\
GRU &Gated recurrent unit\\
HP &Hodrick-Prescott\\
HFCM & High order fuzzy cognitive maps \\
IMF &Intrinsic mode function\\
KDE & Kernel density estimation\\
LLM &Large language model\\
LOESS & Locally estimated scatterplot smoothing\\
LSTM &Long short-term memory\\
MAE &Mean absolute error\\
MAPE &Mean absolute percentage error\\
ML &Machine learning\\
MSTL &Multiple season-trend decomposition using LOESS\\
N-BEATS &Neural basis expansion analysis\\
NHINTS &Neural hierarchical interpolation for time series\\
RMSE &Root-mean-square error\\
RNN &Recurrent neural network\\
SMAPE &Symmetric mean absolute percentage error\\
STGCN &Spatiotemporal graph convolutional network\\
STL &Season-trend decomposition using LOESS\\
SVM &Support vector machine\\
TCN &Temporal convolutional networks\\
TFT &Temporal fusion transformer\\
Time-LLM & Time large language model\\
TPE &Tree-structured Parzen estimator\\
YOLO &You only look once\\
\hline 
\end{tabular}
\end{table}

The proposed hybrid model employs noise attenuation considering an input filter stage. The Christiano Fitzgerald (CF) asymmetric random walk~\citep{dutra2022measuring}, Hodrick-Prescott (HP)~\citep{schuler2024filtering}, season-trend decomposition using LOESS (STL)~\citep{he2021seasonal}, multiple STL (MSTL)~\citep{bandara2025mstl}, empirical wavelet transform (EWT)~\citep{yao2024multi}, Butterworth~\citep{hu2024intelligent}, and empirical mode decomposition (EDM)~\citep{klaar2024random} filters are analyzed, and the best filter is employed in the model input. For the STL and MSTL, the LOESS stands for locally estimated scatterplot smoothing. 

Based on the filtered signal, the LLM is used to perform the time series prediction, also known as timeLLM. To ensure that the best structure is used, the Optuna framework using the tree-structured Parzen estimator (TPE) is considered for hyperparameter tuning \citep{STEFENON2024109876}. Considering the proposed optimized LLM model applied for fault prediction, this paper has the following contributions:

\begin{itemize}
    \item It presents an innovative way of analyzing time series based on LLM models, and is a strategy that can be applied in the future considering the advances in this field.
    \item By using a filter on the input signal, unrepresentative noises are disregarded, making the prediction model more assertive and promising in chaotic time series analysis.
    \item Based on a strategy using TPE, the structure is hyper-adjusted ensuring that the optimal hyperparameters are used in the proposed model.
\end{itemize}

The remainder of this paper is as follows: In section~\ref{2}, related works about fault prediction in insulators are presented. Section~\ref{3} explains the proposed optimized LLM model. In section~\ref{4} the results of the application of the proposed model are discussed, firstly the evaluation of the filter is presented, followed by the tuning strategy. Considering a hypertuned structure, a statistical analysis and benchmarking are presented. Section~\ref{5} presents final remarks and directions for future research. 

\section{Related Works} \label{2}

Fault prediction in distribution grid insulators is critical for maintaining the reliability and safety of power systems~\citep{rocha2019inspection}. Recent studies have explored various methodologies to predict insulator failures, focusing on analyzing leakage currents, employing ML techniques, and utilizing advanced signal processing methods. A way to identify faults in this context involves monitoring the time series data of the leakage current of insulators under contaminated conditions.

Several authors have applied computer vision-based methods considering convolutional neural networks (CNNs)  for insulator fault identification~\citep{9314102}. Prominent in this area is the you only look once (YOLO) model, which was applied in its third generation by~\citet{9866802}, in the fourth generation by~\citet{10012377}, in the fifth generation by~\citet{10106252}, considering a hypertuned version by~\citet{STEFENON2024102722}, or using a hybrid version in~\citep{gtd2.12886}. 

\citet{deng2022research} proposed a modified YOLO that can be computed on edge devices. The modification to make the algorithm more efficient was in the backbone of the YOLO. Instead of using a CSPDarknet53 (standard for the YOLO version that they considered), they applied a lightweight network MobilieNetv3. Based on this modification, the proposed algorithm achieved an accuracy of 0.945 with a speed of 58.5 frames per second. \citet{10298815} also applied a modified version of YOLO, considering a model called multi-fault insulator detection they achieved an accuracy of 0.939.

\citet{9537742} presented an improved faster region CNN for insulator fault detection. They considered preprocessing techniques for image segmentation, reducing the image noise, and having the focus on the insulators, even when complex backgrounds are considered. Based on their model, it was possible to achieve a mean average precision of 0.908 considering glass insulators and 0.917 when composite insulators were evaluated. Based on CNNs, \citet{10258031} considered infrared images for insulator defect diagnosis. Applications based on CNNs are promising, as they can be handled indirectly (without contact with the electrical network)~\citep{singh2023interpretable}, while in the case of leakage current it is necessary to measure the insulators directly, making it a more challenging approach.

\citet{s22166121} subjected insulators to saline environments to simulate contamination and analyzed the progression of leakage currents leading up to disruptive discharges. The researchers evaluated several time series forecasting models, including group method of data handling (GMDH), long short-term memory (LSTM), adaptive neuro-fuzzy inference system (ANFIS), and various ensemble learning models. They found that integrating wavelet transforms with these models improved prediction accuracy, with the wavelet-ANFIS model achieving the best performance. In \citep{s23136118} the Christiano-Fitzgerald filter was combined with the GMDH to predict faults in contaminated insulators.

Studies of the contamination process that leads to the development of flashover, such as the one presented by~\citet{9828523} are rarer. Especially when it comes to applications of ML models, as presented by~\citet{10587253}, or hybrid methods for time series forecasting. In other applications, some authors have researched how to better identify faults based on time series analysis, as presented in Table~\ref{re_w}.

\begin{table}[]
%\footnotesize
\small
\caption{Fault prediction and anomaly detection using ML approaches considering time series.}
\label{re_w}
\centering
\begin{tabular}{p{3.6cm}p{6cm}p{5cm}}\hline
Work  & Method & Application\\ \hline
\citep{10356858} & Parallel time series modeling with LightNet and DarkNet. & Fault detection on intelligent vehicles.\\
\citep{guo2021mechanical} & Error fusion of multiple sparse auto-encoder LSTM. & Mechanical fault prediction.\\
\citep{xu2021two} & Attention-based-LSTM, random forest, and extra-tree. & Machinery fault prediction.
\\
\citep{liu2021machinery} & CNN, gated recurrent unit (GRU), attention, and knowledge graph. & Machinery fault diagnosis.\\
\citep{hsu2021multiple} & Multiple time-series CNN. & Fault %detection and diagnosis 
in semiconductor. \\% manufacturing.\\
\citep{he2022graph} & Masked spatial graph attention network with GRU. & Fault detection for unmanned aerial vehicles.\\
\citep{jin2022time} & Time series transformer. & Machinery fault diagnosis. \\
\citep{chen2023unsupervised} & Unsupervised deep autoencoder with dimension fusion function. & Fault detection in aeroengines. \\
\citep{zhang2023gaussian} & Gaussian-linearized transformer with %tranquilized time series 
decomposition. & Fault diagnosis in methane gas sensors.\\
\citep{xie2021attention} & CNN-LSTM with attention. & Wind turbine fault prediction.\\
\citep{arunthavanathan2021deep} & CNN-LSTM with one-class support vector machine (SVM). & Fault detection in multivariate complex process systems.\\
\citep{nguyen2021forecasting} & 
Standard LSTM and an LSTM autoencoder with a one-class SVM. & Anomaly detection in supply chain management.\\
\citep{9612196} & Spatial and temporal attention-based GRU with seasonal-trend decomposition. & Fault diagnosis  of electro-mechanical actuators.\\
\citep{10662411} & Autoformer enhanced by Dilated loss module. & Potential bushing and transformer faults.\\
\citep{chen2021bearing} & Multi-scale CNN and LSTM. & Bearing fault diagnosis. \\
\citep{s22218323} & Wavelet tranform with LSTM. & Fault %forecasting 
in power grids.\\
\citep{211126} & Linear regression, support vector regression, multilayer Perceptron, deep neural network, and RNNs. & Failure prediction in contaminated insulators.\\
\citep{SEMAN2023109269} & Ensemble random subspace with Hodrick–Prescott filter.& Fault forecasting in pin-type insulators.\\
\citep{klaar2023optimized} & EWT-sequence-to-sequence-LSTM with attention mechanism. &  Insulators fault prediction.\\
\citep{3076410} & Stacking ensemble learning model with wavelet transform. & Insulators contamination forecasting.\\
\citep{branco2024bootstrap} & Bootstrap aggregation with Christiano–Fitzgerald random walk filter. & Fault prediction based on leakage current.\\
\citep{en13020484} & Adaptive neuro-fuzzy inference system with wavelet packets transform. & Insulator fault forecasting.\\
\hline 
\end{tabular}
\end{table}

Several authors (see Table \ref{re_w}) presented hybrid methods for predicting time series, such as CNN-LSTM, which uses CNN for feature extraction and LSTM for time series forecasting, showing that combining techniques with different objectives can improve the architecture, resulting in a hybrid approach that outperforms the lasted architectures for time series forecasting. A more detailed discussion of which state-of-the-art models are used for time series forecasting is presented in the next subsection.

\subsection{State-of-the-Art in Time Series Forecasting}

In recent years, advances in ML and DL models have driven the development of more robust and accurate methods to predict time series. 
In this regard, DL-based approaches including recurrent neural networks (RNN)~\citep{rnnTS2021} and transformers~\citep{transformerTS2023}, have shown promising performance.
In~\citep{tft2021} the authors considered the use of the temporal fusion transformer (TFT) model for time series. This model is an attention-based deep neural network architecture designed for multi-horizon time series forecasting, combining high performance with interpretability. The model incorporates static covariate encoders, gating mechanisms, variable selection networks, and hybrid temporal processing, which uses LSTMs for local patterns and self-attention to capture long-term dependencies.

\citet{NBEATS2020} proposed the neural basis expansion analysis for time series forecasting (N-BEATS), which is a deep neural network architecture designed for univariate time series forecasting based on residual connections and multiple fully connected layers. The deep temporal convolutional network (DeepTCN) was proposed in~\citep{cnnTS2020} and is a CNN architecture developed for probabilistic forecasting of multiple related time series. The model employs dilated causal convolutions, which guarantee dependence only on past inputs and capture long-range patterns with computational efficiency. DeepTCN demonstrates robustness outperforming models such as seasonal autoregressive integrated moving average, light gradient boosting machine~\citep{lightgbmTS2017}, and the probabilistic forecasting with autoregressive recurrent network~\citep{deepARTS2020}.

In~\citep{dotsTS2020} the authors presented the multivariate time series forecasting with a graph neural networks model. This is a graph neural network designed-based model to forecast multivariate time series by learning the underlying graph structure. Its architecture combines three main components: a graph learning layer, which extracts dynamic relationships between series without the need for a predefined graph structure; graph convolution modules, which model spatial dependencies between variables; and temporal convolution modules, which capture long-term patterns through dilated convolutions, making it a robust approach for spatiotemporal forecasting in different datasets types.

\citet{spatioTS2018} proposed the spatiotemporal graph convolutional network (STGCN), which is a DL architecture designed for traffic prediction, combining graph convolutions and temporal convolutions with gated linear units to efficiently model spatial and temporal dependencies. The architecture is composed of spatiotemporal convolutional blocks, where causal temporal convolutions extract sequential patterns and graph convolutions capture spatial relationships without relying on a fixed grid structure. STGCN is computationally efficient, enabling fast and scalable training for large networks. The model outperformed other approaches such as 
full-connected LSTM and graph convolutional GRU in several error metrics such as mean absolute error (MAE), mean absolute percentage error (MAPE), and root-mean-square error (RMSE), consuming up to 14 times less training time compared to state-of-the-art models.

In~\citep{hybridTS2023} the authors claim that using hybrid models can improve time series forecasting. The hybrid model proposed by the authors combines CNNs, attention-based LSTM (A-LSTM), and an auto-regressive model (AR) to forecast energy generation from multiple renewable sources. CNN captures spatial correlations between energy sources, A-LSTM models non-linear temporal patterns, and AR extracts linear trends. Its main advantages include the ability to model capture complex temporal patterns and predictive superiority over traditional models such as artificial neural networks and decision trees. The model also demonstrated a significant reduction in prediction errors compared to previous state-of-the-art approaches, reducing MAE by up to 27.1\% and MAPE by 53.6\% for photovoltaic solar energy. The high R$^{2}$ values ($>$0.945) confirm its good fit with the observed data, making it a robust and effective solution for forecasting renewable energy.

\citet{emWavTS2023} employed the EWT high-order fuzzy cognitive map that is also a hybrid model for forecasting time series that combines the EWT, high-order fuzzy cognitive maps (HFCM), and ridge regression. The model's architecture decomposes the time series data with EWT, which adapts filters to the signal spectrum, followed by modeling with HFCM, which captures long-term temporal dependencies, and hyperparameter optimization with ridge regression, avoiding overfitting. The forecast is reconstructed using inverse EWT, allowing complex patterns in non-stationary series to be captured. Evaluated with RMSE on 15 real data sets, the model outperformed 11 state-of-the-art algorithms, including LSTM, RNN, ANFIS, temporal convolutional network (TCN), and CNN-fuzzy cognitive maps.

~\citet{timellmTS2024} proposed a time series forecasting by reprogramming large language models (time-LLM) framework to predict time series, without modifying the backbone language model. The approach transforms time series into prototypical textual representations and uses the prompt-as-prefix technique to improve the input context. Its main advantages include generalization to multiple domains, data efficiency, advanced reasoning capabilities, and no need for fine-tuning, allowing robust predictions even with few examples or in zero-shot learning scenarios. Evaluated on different datasets, time-LLM outperformed conventional models in metrics such as mean-square error, MAE, symmetric MAPE (SMAPE), and overall weighted average, demonstrating high accuracy in short and long-term forecasts.

In~\citep{hyperparameterTS2024} proposes the use of CNN-LSTM for wind energy forecasting, with advanced hyperparameter optimization to improve accuracy and efficiency. CNN is used to extract spatial patterns from the data, while LSTM models short- and long-term temporal dependencies. Different hyperparameter optimization algorithms are evaluated, %(Scikit-opt, Optuna, and Hyperopt), 
with Bayesian optimization via TPE being the most common approach. The results show that the advanced selection of hyperparameters significantly improves the effectiveness of wind energy forecasting, making the models more %accurate and 
reliable.

These studies underscore the importance of integrating advanced analytical methods, such as ML and signal processing, with traditional monitoring techniques to improve fault prediction in distribution grid insulators. Continued research in this area is essential to develop more accurate and reliable predictive models, ultimately contributing to the improved stability and efficiency of power distribution grids.

\section{Methodology} \label{3}

This paper considers the use of a filter input stage for noise attenuation and a hypertuned LLM model for time series forecasting. The time series is denoised to remove high frequencies and to have the focus of the analysis on the variation trend. To ensure that an optimal structure is assumed, the hyperparameters of the LLM are tuned via multi-criteria optimization based on TPE using the Optuna framework \citep{srinivas2022hyoptxg}. In this section, all considered techniques employed in the proposed optimized LLM model are explained.

\subsection{Input Stage Filter}

Filters are applied in this paper to reduce the high frequencies in the signal, thus focusing on predicting the trend \citep{ouyang2021stl}, which represents the most significant variation in the temporal analysis that leads to failure after a certain time in conditions of high contamination.
The focus of the prediction here is the trend because its variation represents the increase in the leakage current until it reaches the limit accepted by the insulator.

The CF, HP, STL, MSTL, EWT, Butterworth, and EDM filters are analyzed for signal denoising, and the best filter is employed in the model input stage. %These filters are explained in the following. 
The symbols used in the equation of these filters are presented in Table \ref{table_sym}. The symbols that are used in both the filters and the model are presented in the section that explains the model architecture.

\begin{table}[ht]
%\footnotesize
\small
\begin{tabular}{llcll}\hline
Symbol & Definition\\ \hline

$a$, $b$ & Coefficients of the filter\\
$c$ & Intrinsic mode function \\
$e$& Envelop of the IMFs \\
$h$ & Candidate IMF\\
$m$ & Envelopes of envelopes\\
$n$ & Order of the filter\\
$p$ & Lower filter window \\
$q$ & Upper filter window \\
$s$ & Seasonal component\\
$w$ & Filter weights \\

\hline
$F$& Fourier transform\\
$H$ & Transfer function \\
$M$ & Number of seasonal components \\
$N$ & Number of modes of decomposition\\
$P$ & Degree of the polynomial\\
$S$ & Laplace transform variable\\
$T$ & Sampling period\\
$W$ & Empirical wavelet component\\
$Z$ & Complex variable in the $z$-domain\\

\hline
$\beta$ & Coefficient of the polynomial \\
$\delta$ & Width of the transition band\\
$\epsilon$ & Residual component\\
$\kappa $ & Cutoff frequency& \\
$\lambda$ & Smoothing hyperparameter \\

$\tau$ & Trend component\\
$\phi$ & Scaling function \\
$\psi$ & Wavelet function\\
$\omega$ & Angular frequency\\

%& Time step & $t$\\
 
%each segment $[\omega_{k-1}, \omega_k]$, a 

%& $\mathbf{x}_t$ & vector of predictor variables \\
%& $\mathbf{X}$ & design matrix for the local neighborhood \\
%& $\mathbf{W}_t$ & diagonal weight matrix \\ 
%& $\mathbf{y}$ & vector of observed data points \\
\hline
\end{tabular} \centering
\caption{Symbols used in filter equations.}
\label{table_sym}
\end{table}

\subsubsection{Christiano Fitzgerald Asymmetric Random Walk (CF) Filter}

The CF filter is designed for the decomposition of time series data into seasonal and trend components. Unlike symmetric filters, the CF filter can operate asymmetrically, making it useful for real-time forecasting.
The CF filter is based on the concept of band-pass filtering, where the goal is to isolate cycles within a specified frequency range~\citep{bhowmik2021cyclical}. The mathematical foundation of the filter involves approximating the ideal band-pass filter, which has a transfer function defined as:
\begin{align}
H(\omega) &= \begin{cases} 
1 & \text{if } \kappa_{\text{lower}} \leq |\omega| \leq \kappa_{\text{upper}}, \\
0 & \text{otherwise},
\end{cases}
\end{align}
where $\kappa_{\text{lower}}$ and $\kappa_{\text{upper}}$ are the lower and upper cutoff frequencies, respectively. 
The filter is implemented as:
\begin{align}
\hat{s}_t &= \sum_{j=-p}^q w_j y_{t-j},
\end{align}
where $y_t$ is the observed time series, $\hat{s}_t$ is the estimated seasonal component, $w_j$ are the filter weights, and $p$ and $q$ define the range of the filter window. The weights $w_j$ are determined by minimizing the error between the filtered signal and the ideal band-pass filter. Mathematically, this involves solving:
\begin{align}
\min_{w_j} \int_{-\pi}^{\pi} \left| H(\omega) - \sum_{j=-p}^q w_j e^{-i \omega j} \right|^2 d\omega.
\end{align}

When future observations are not available, necessitating the use of an asymmetric filter, the CF filter solves this by adjusting the weights such that:
\begin{align}
\hat{s}_t &= \sum_{j=0}^q w_j y_{t-j},
\end{align}
where only past and present observations are used. The computation of these asymmetric weights involves a recursive approach to approximate the frequency response of the ideal band-pass filter. The CF filter provides a robust method for decomposing and forecasting time series data in real-time applications~\citep{s23136118}. 

\subsubsection{Hodrick-Prescott (HP) Filter}

The HP filter is used in time series analysis to decompose a series into its trend and seasonal components. It is particularly for extracting the smooth long-term trend from noisy data~\citep{li2023multi}. 
The HP filter decomposes a time series $y_t$ into trend $\tau_t$ and seasonal $s_t$ components.
The estimation of $\tau_t$ is formulated as a minimization problem. Specifically, the objective function combines a goodness-of-fit term and a smoothness penalty:
\begin{align}
\min_{\{\tau_t\}} \sum_{t=1}^T (y_t - \tau_t)^2 + \lambda \sum_{t=2}^{T-1} \left[ (\tau_{t+1} - \tau_t) - (\tau_t - \tau_{t-1}) \right]^2.
\end{align}
where the first term, $\sum_{t=1}^T (y_t - \tau_t)^2$, ensures that the estimated trend $\tau_t$ fits the data well. The second term, $\sum_{t=2}^{T-1} \left[ (\tau_{t+1} - \tau_t) - (\tau_t - \tau_{t-1}) \right]^2$, penalizes changes in the slope of the trend, effectively ensuring that $\tau_t$ is smooth over time. The smoothing hyperparameter $\lambda$ controls the trade-off between these two objectives. 

The choice of $\lambda$ is subjective and may affect the decomposition. This issue can be solved with a $\lambda$ adjustment. When $\lambda \to 0$, the trend $\tau_t$ closely follows the original series $y_t$, allowing for greater flexibility, on the other hand, when $\lambda \to \infty$, the trend becomes a linear function, as the penalty on deviations from linearity dominates~\citep{sakarya2022spectral}.

%Common values of $\lambda$ depend on the frequency of the data. For example, $\lambda = 100$ is used for annual data, $\lambda = 1600$ for quarterly data, and $\lambda = 14400$ for monthly data.

%While the HP filter is a powerful tool, it has limitations. The choice of $\lambda$ is subjective and can significantly affect results. The filter also assumes a smooth trend, which may not be appropriate for all data types. Moreover, it can introduce spurious dynamics near the endpoints of the series, a phenomenon known as end-point bias.

\subsubsection{Season-Trend Decomposition using LOESS (STL) Filter}

The operation in STL decomposition is based on the application of LOESS, a regression method used to estimate a smooth function by fitting weighted local polynomials~\citep{xu2023offshore}. The LOESS procedure at any time point $t$ involves: Initially, it defines the neighborhood around $t$ by selecting points within a specified window determined by the smoothing hyperparameter $\lambda$, which controls the fraction of data used in the local fit.
Thus, the weights are assigned to the observations within the neighborhood based on their distances from $t$ using a kernel function, commonly the tricube kernel:
\begin{align}
w(x) = (1 - |x|^3)^3, \quad \text{for } |x| \leq 1, \quad w(x) = 0, \text{otherwise}.
\end{align}

Fitting a weighted polynomial (typically linear or quadratic) to the data within the neighborhood, minimizing the locally weighted sum of squared residuals, given by
\begin{align}
\min_{\beta_0, \beta_1, \dots, \beta_P} \sum_{i} w(x_i) \left( y_i - \sum_{j=0}^{P} \beta_j x_i^j \right)^2,
\end{align}
where $\beta_0, \beta_1, \dots, \beta_P$ are the coefficients of the polynomial, and $P$ is the degree of the polynomial (commonly $P = 1$ or $P = 2$). Once the decomposition is complete, the trend component $\tau_t$ can be used for forecasting. For example, future values of $y_t$ may be predicted by evaluating the trend $\tau_t$ using prediction models. The remainders are often modeled as a stochastic noise, assumed as white noise~\citep{klaar2023structure}.

\subsubsection{Multiple Season-Trend Decomposition using LOESS (MSTL) Filter}

The MSTL is a method used for analyzing time series data, particularly when multiple seasonal components and trends are present. The primary goal of MSTL is to decompose the time series into trend, seasonality, and remainder components according to Equation~\ref{eq_base}. Mathematically, for a univariate time series $y_t$, the method can be represented as follows:
\begin{align}
y_t = \tau_t + \sum_{j=1}^M s_{j,t} + \epsilon_t,
\end{align}
where $s_{j,t}$ is the seasonal component for the $j$-th seasonal frequency, where $j = 1, \ldots, M$, $M$ is the number of seasonal components, and $\epsilon_t$ is the residual component~\citep{sohrabbeig2023decompose}.

Like the STL, the MSTL employs the concept of LOESS to estimate each component iteratively. LOESS operates by fitting a polynomial regression model locally for a subset of data points around each time index $t$, weighted by a kernel function~\citep{rhif2022detection}. The kernel assigns higher weights to points closer to $t$ and lower weights to points farther away. 
%The LOESS estimator for a single point is expressed as:
%\begin{align}
%\hat{y}_t = \mathbf{x}_t^\top (\mathbf{X}^\top \mathbf{W}_t \mathbf{X})^{-1} \mathbf{X}^\top \mathbf{W}_t \mathbf{y},
%\end{align}
%where $\mathbf{x}_t$ is the vector of predictor variables at time $t$, $\mathbf{X}$ is the design matrix for the local neighborhood, $\mathbf{W}_t$ is a diagonal weight matrix determined by the kernel function, and $\mathbf{y}$ is the vector of observed data points.

In the MSTL decomposition process, the seasonal components $s_{j,t}$ are estimated first. Each seasonal component corresponds to a specific periodicity, and LOESS smoothing is applied after aggregating data for that periodicity. MSTL is robust to outliers through the use of LOESS, where weights are iteratively adjusted based on residuals to reduce the influence of extreme values~\citep{elseidi2024mstl}.

\subsubsection{Empirical Wavelet Transform (EWT) Filter}

The EWT is a signal decomposition method that is designed to extract meaningful frequency components from a signal. Unlike traditional wavelet transforms, which rely on predefined mother wavelets and fixed frequency partitions, the EWT constructs wavelet filters based on the spectral characteristics of the input signal. This adaptability makes it a promising method for time series forecasting, especially in cases where signals exhibit non-stationary behavior~\citep{mohammadi2023using}.

For each segment $[\omega_{k-1}, \omega_k]$, a scaling function $\phi_k(\omega)$ and a wavelet function $\psi_k(\omega)$ are constructed. These functions are designed to satisfy orthogonality and completeness conditions over the defined frequency intervals. The $\phi_k(\omega)$ is used to capture the low-frequency components, while the $\psi_k(\omega)$ isolates the band-limited frequency components~\citep{peng2022effective}. The empirical scaling function and wavelet function in the Fourier domain can be defined as:

\begin{align}
\phi_k(\omega) = 
\begin{cases} 
1, & \omega \in [0, \omega_k - \delta_k], \\
\cos\left(\frac{\pi}{2\delta_k} (\omega - (\omega_k - \delta_k))\right), & \omega \in [\omega_k - \delta_k, \omega_k], \\
0, & \text{otherwise},
\end{cases}
\end{align}

\begin{align}
\psi_k(\omega) = 
\begin{cases} 
1, & \omega \in [\omega_{k-1} + \delta_k, \omega_k - \delta_k], \\
\sin\left(\frac{\pi}{2\delta_k} (\omega - (\omega_{k-1} + \delta_k))\right), & \omega \in [\omega_{k-1}, \omega_{k-1} + \delta_k], \\
\cos\left(\frac{\pi}{2\delta_k} (\omega - (\omega_k - \delta_k))\right), & \omega \in [\omega_k - \delta_k, \omega_k], \\
0, & \text{otherwise},
\end{cases}
\end{align}
where $\delta_k$ is the width of the transition band. 

After constructing the wavelet and scaling functions, the signal $x(t)$ is decomposed into empirical wavelet coefficients using the inverse Fourier transform. The $k$th empirical wavelet component is given by:

\begin{align}
W_k(t) = F^{-1} \left[ \hat{x}(\omega) \psi_k(\omega) \right],
\end{align}

\noindent where $F^{-1}$ denotes the inverse Fourier transform. Similarly, the residual low-frequency component is captured by:

\begin{align}
s_N(t) = F^{-1} \left[ \hat{x}(\omega) \phi_N(\omega) \right].
\end{align}

Thus, the original signal can be reconstructed as the sum of the decomposed components:

\begin{align}
x(t) = s_N(t) + \sum_{k=1}^{N} W_k(t).
\end{align}

The EWT is particularly advantageous for time series forecasting as it effectively isolates the dominant modes of variability, allowing for independent modeling of each component. This modularity facilitates the application of predictive models, such as prediction models, neural networks, or hybrid approaches, on each extracted component~\citep{8769843}.

\subsubsection{Butterworth Filter}

The Butterworth filter is a signal processing technique used in time series analysis and forecasting due to its smooth frequency response and minimal distortion characteristics. It effectively separates high-frequency noise from low-frequency trends in time series data. 
The Butterworth filter is applied to have a maximally flat magnitude response in the passband, avoiding ripples in the frequency response~\citep{hu2024intelligent}. The general transfer function for an $n$th-order Butterworth filter is given by:

\begin{align}
H(\omega) = \frac{1}{\sqrt{1 + \left(\frac{\omega}{\kappa}\right)^{2n}}},
\end{align}

\noindent where $\omega$ is the angular frequency, $\kappa$ is the cutoff frequency, and $n$ is the order of the filter. The cutoff frequency $\kappa$ defines the boundary between the passband and the stopband. The filter’s order, $n$, defines the steepness of the transition between the passband and the stopband. Higher-order filters yield sharper transitions but introduce higher computational complexity~\citep{9445054}.

%From a mathematical perspective, the Butterworth filter’s design involves solving the minimization problem to achieve the flattest magnitude response in the passband. The poles of the transfer function are derived by equating the denominator of $H(s)$ to zero:

%\begin{align}
%1 + \left(\frac{\omega}{\kappa}\right)^{2n} = 0.
%\end{align}

%This results in $2n$ poles symmetrically distributed on the unit circle in the $s$-domain, ensuring stability and causal behavior when implemented in digital systems.

In time series forecasting, the Butterworth filter is often used as a low-pass filter to extract the low-frequency trend component of a time series. This is accomplished by suppressing high-frequency variations, which are typically associated with noise or short-term fluctuations while retaining the underlying trend~\citep{singh2023novel}.
To apply the Butterworth filter to discrete time series data, the transfer function is transformed into the $Z$-domain using the bilinear transformation:

\begin{align}
S =  \left (\frac{2}{T}  \right ) \frac{1 - Z^{-1}}{1 + Z^{-1}},
\end{align}

\noindent where $S$ is the Laplace transform variable, $T$ is the sampling period, and $Z$ is the complex variable in the $Z$-domain. Substituting this transformation into the continuous-time transfer function yields the discrete-time transfer function:

\begin{align}
H(Z) = \frac{\sum_{k=0}^{n} b_k Z^{-k}}{1 + \sum_{k=1}^{n} a_k Z^{-k}},
\end{align}

\noindent where the coefficients $a_k$ and $b_k$ are determined by the order of the filter and the cutoff frequency. The recursive form of this difference equation enables efficient implementation of the filter in time series forecasting applications \citep{guodong2023comparison}.

The Butterworth filter minimizes distortion in the trend component by ensuring that the magnitude response in the passband is as flat as possible. %Although the filter is not inherently linear-phase, which can introduce phase distortion in the signal, this issue can be mitigated by applying the filter in a forward and backward direction. %This approach, known as zero-phase filtering, eliminates phase distortion by combining the forward and backward passes:
%
%\begin{align}
%x_t^{\text{filtered}} = \text{filt}(x_t),
%\end{align}
%
%\noindent where $\text{filt}$ denotes the zero-phase filtering operation.
%
This flexibility in adjusting the cutoff frequency and filter order allows it to be tailored to the specific characteristics of the time series under study, making it a powerful method for preprocessing and smoothing data before modeling and prediction~\citep{chawuthai2022travel}.

\subsubsection{Empirical Mode Decomposition (EMD) Filter}

The EMD is an adaptive, data-driven technique for analyzing non-linear and non-stationary time series data. It decomposes a signal into a finite set of components called intrinsic mode functions (IMFs) and a residual. The decomposition process begins by identifying the IMFs, which are defined by two conditions: the number of extrema and zero crossings in an IMF must either be equal or differ at most by one, and the mean value of the upper envelope and the lower envelope must be zero at every point \citep{ying2021permutation}.

To formalize the process, consider a time series $x(t)$. The first step involves constructing envelopes for the signal using cubic splines to interpolate the local maxima and minima. Denoting the upper and lower envelopes as $e_{\text{upper}}(t)$ and $e_{\text{lower}}(t)$, respectively, their mean $m(t)$ is calculated as
\begin{align}
m(t) = \frac{e_{\text{upper}}(t) + e_{\text{lower}}(t)}{2}.
\end{align}

The mean is then subtracted from the original signal to produce a candidate IMF:
\begin{align}
h(t) = x(t) - m(t).
\end{align}

This process, known as sifting, is iterated until $h(t)$ satisfies the IMF criteria. The first IMF ($c_1(t)$), captures the highest frequency oscillations in the signal. To extract subsequent IMFs, $c_1(t)$ is subtracted from the original signal:
\begin{align}
\epsilon_1(t) = x(t) - c_1(t),
\end{align}
where $\epsilon_1(t)$ becomes the new input signal for further decomposition. Repeating this procedure yields a set of IMFs $\{c_1(t), c_2(t), \dots, c_n(t)\}$ and a final residual $\epsilon_n(t)$, such that
\begin{align}
x(t) = \sum_{i=1}^n c_i(t) + \epsilon_n(t).
\end{align}

Each IMF isolates oscillatory modes of different scales, making EMD particularly suitable for analyzing signals where distinct temporal scales are present.
For time series forecasting, consider that the extracted IMFs often exhibit simpler patterns than the original signal. These components can then be modeled independently using ML methods. Forecasting can proceed by predicting each IMF $c_i(t)$ individually by ML models~\citep{liu2021application}. 

\subsection{Prediction Model Architecture}

LLM applied for time series, or timeLLM, is built upon the foundational structure of large language models by incorporating temporal reasoning directly into its architecture. The innovation of this model lies in the integration of temporal embeddings, dynamic contextual adjustments, and attention mechanisms for time-dependent data. These features enhance the model's ability to handle sequential and time-sensitive information~\citep{10856008}.
The symbols used in the equation of the model architecture and its hypertuning are presented in Table \ref{table_sym2}.

\begin{table}[ht]
%\footnotesize
\small
\begin{tabular}{llcll}\hline
Symbol & Definition\\ \hline
$d$ & Embedding dimension\\
$i$, $k$ & Iteration \\
$l$, $g$ & Conditional densities \\
$p$ & Probability\\
$t$ & Timestamps or step \\
$x$ & Input data\\
$y$ & Observed time series (objective)\\
\hline
$\gamma$ & Quantile of the observed objective \\
$\eta$ & Event in a sequence\\
$\mathcal{G}$ & Gating function \\
$\mathcal{K}$ & Kernel function\\
$\mathcal{L}$ & Loss function\\
$\mathcal{N}$ & Normalization constant\\
$\mathcal{X}$ & Hyperparameter space \\
\hline
$\mathbf{h}$ & Modified input representation\\
$\mathbf{v}$ & Vector representation \\
$\mathbf{E}$ & Standard token embedding\\ 
$\mathbf{K}$& Key matrix\\
$\mathbf{Q}$ & Query matrix\\
$\mathbf{T}$ & Temporal embedding\\
$\mathbf{V}$& Value matrix\\
$\mathbf{W}$ & Time-aware weighting matrix\\
\hline
\end{tabular} \centering
\caption{Symbols used in model equations.}
\label{table_sym2}
\end{table}

Temporal embeddings encode time-related information, such as timestamps or relative durations, by mapping these into a latent space. Mathematically, given a sequence of events 
$\{\eta_1, \eta_2, \dots, \eta_n\}$ occurring at corresponding timestamps $\{t_1, t_2, \dots, t_n\}$, temporal embeddings $\mathbf{T}(t)$ map each timestamp $t_i$ to a vector representation $\mathbf{v}_i \in \mathbb{R}^d$, where $d$ is the embedding dimension. These embeddings are learned jointly with the model hyperparameters, ensuring that temporal context is captured alongside linguistic features~\citep{10771272}.

The temporal embeddings are integrated into the model's transformer layers. For each input token $x_i$, the modified input representation is given by
\begin{align}
\mathbf{h}_i = \mathbf{E}(x_i) + \mathbf{T}(t_i),
\end{align}
where $\mathbf{E}(x_i)$ is the standard token embedding and $\mathbf{T}(t_i)$ is the temporal embedding. This addition ensures that the model incorporates temporal information at the input stage.

The attention mechanism in timeLLM is adapted to prioritize temporal dependencies. The scaled dot-product attention is modified to include a temporal weighting term. For query $\mathbf{Q}$, key $\mathbf{K}$, and value $\mathbf{V}$ matrices, the attention weights are computed as
\begin{align}
\text{Attention}(\mathbf{Q}, \mathbf{K}, \mathbf{V}) = \text{softmax}\left( \frac{\mathbf{Q}\mathbf{K}^T}{\sqrt{d_k}} + \mathbf{W}_T \right) \mathbf{V},
\end{align}
where $\mathbf{W}_T$ is a time-aware weighting matrix that adjusts the attention based on temporal proximity~\citep{jin2023time}.
Incorporating dynamic contextual awareness, timeLLM leverages time-dependent positional encodings and gating mechanisms. The model introduces a gating function $\mathcal{G}(t)$ that modulates the influence of temporal embeddings based on the recency or relevance of information. The gated representation is computed as
\begin{align}
\mathbf{h}_i' = \mathcal{G}(t_i) \cdot \mathbf{h}_i,
\end{align}
where $\mathcal{G}(t_i)$ is a learnable function of $t_i$.

TimeLLM has practical applications in event forecasting, where it analyzes historical data to predict future outcomes. For example, given a time series $\{(t_i, x_i)\}$ representing observed data points, the model learns to predict $x_{n+1}$ by optimizing the loss function
\begin{align}
\mathcal{L} = \sum_{i=1}^n \left( x_i - \hat{x}_i \right)^2 + \lambda \| \mathbf{T}(t) \|^2,
\end{align}
where the second term regularizes the temporal embeddings to prevent overfitting.
By analyzing cross-attention patterns between temporally ordered inputs, timeLLM identifies dependencies that inform predictions and interpretations~\citep{10621076}. The incorporation of these mechanisms makes timeLLM a robust solution for time-sensitive applications, such as time series forecasting presented in this paper.

\subsection{Model Hypertuning}

For hyperparameter model tuning, the TPE is applied. TPE is a model-based optimization algorithm that builds on the Bayesian optimization, tailored for hyperparameter tuning \citep{zhang2024carbon}. The TPE algorithm uses probabilistic modeling to construct surrogate models of the objective function and guides the search for optimal hyperparameter configurations. In this paper, the TPE is applied for hypertuning the proposed optimized LLM model, the considered hyperparameters are the batch size, dropout, learning rate, and number of heads. 

The goal of hyperparameter optimization is to minimize or maximize an objective function $f: \mathcal{X} \to \mathbb{R}$, where $\mathcal{X}$ is the hyperparameter space \citep{STEFENON2024133918}. 
The TPE idea relies on modeling the conditional probability $p(y \mid x)$, where $x \in \mathcal{X}$ is a vector of hyperparameters and $y = f(x)$ is the corresponding objective value. 
The TPE replace $p(y \mid x)$ with two conditional densities, $l(x)$ and $g(x)$, based on a threshold $\gamma$ such that:
\begin{align}
l(x) = p(x \mid y < \gamma), \quad g(x) = p(x \mid y \geq \gamma),
\end{align}
where $\gamma$ is a quantile of the observed objective values $\{y_1, y_2, \dots, y_n\}$, often chosen as a fixed percentile. The TPE algorithm thus rewrites the marginal likelihood $p(x \mid y)$ using Bayes' theorem:
\begin{align}
p(x \mid y) = 
\begin{cases} 
l(x) p(y < \gamma)/\mathcal{N} & \text{if } y < \gamma, \\ 
g(x) p(y \geq \gamma)/\mathcal{N} & \text{if } y \geq \gamma,
\end{cases}
\end{align}
where $\mathcal{N} = p(y)$ is the normalization constant.

TPE focuses on maximizing the expected improvement (EI) by choosing hyperparameter configurations that are likely to improve upon the current best observations. The EI criterion is reformulated in TPE by using the ratio of densities $l(x)/g(x)$. Intuitively, TPE selects $x$ to maximize this ratio, favoring regions of the search space that are more probable under $l(x)$ (regions associated with good performance) while being less probable under $g(x)$ (regions associated with worse performance)~\citep{tuleski2024audio}.

The algorithm constructs $l(x)$ and $g(x)$ using kernel density estimation (KDE). For a set of observed hyperparameter configurations $\{x_1, x_2, \dots, x_n\}$ and their corresponding objective values $\{y_1, y_2, \dots, y_n\}$, the densities are modeled as:
\begin{align}
l(x) = \sum_{i \in \mathcal{I}_l} w_i \mathcal{K}(x, x_i), \quad
g(x) = \sum_{i \in \mathcal{I}_g} w_i \mathcal{K}(x, x_i),
\end{align}
where $\mathcal{I}_l = \{i \mid y_i < \gamma\}$, $\mathcal{I}_g = \{i \mid y_i \geq \gamma\}$, $w_i$ are the weights associated with each sample, and $\mathcal{K}(x, x_i)$ is a kernel function. The threshold $\gamma$ is updated dynamically as new observations are added to the dataset.

By modeling $l(x)$ and $g(x)$ using KDE, TPE can efficiently handle complex, multimodal distributions over the hyperparameter space. Moreover, it exploits the tree-structured dependencies within $x$ by breaking the optimization problem into sub-problems, each model with a separate density estimator.
The hyperparameter configuration for the next evaluation is sampled by optimizing the acquisition function, which in TPE is the maximization of the density ratio:
\begin{align}
x^* = \arg\max_{x \in \mathcal{X}} \frac{l(x)}{g(x)}.
\end{align}

This process is repeated iteratively, with each new observation refining the density estimators $l(x)$ and $g(x)$, guiding the search toward better regions of the hyperparameter space. TPE is particularly effective for problems with hierarchical or conditional hyperparameter structures, making it well-suited for modern ML models~\citep{DASILVA2024110275}. 

\begin{figure*}[htb!]
	\centering
	\setkeys{Gin}{width=1\textwidth}{\includegraphics{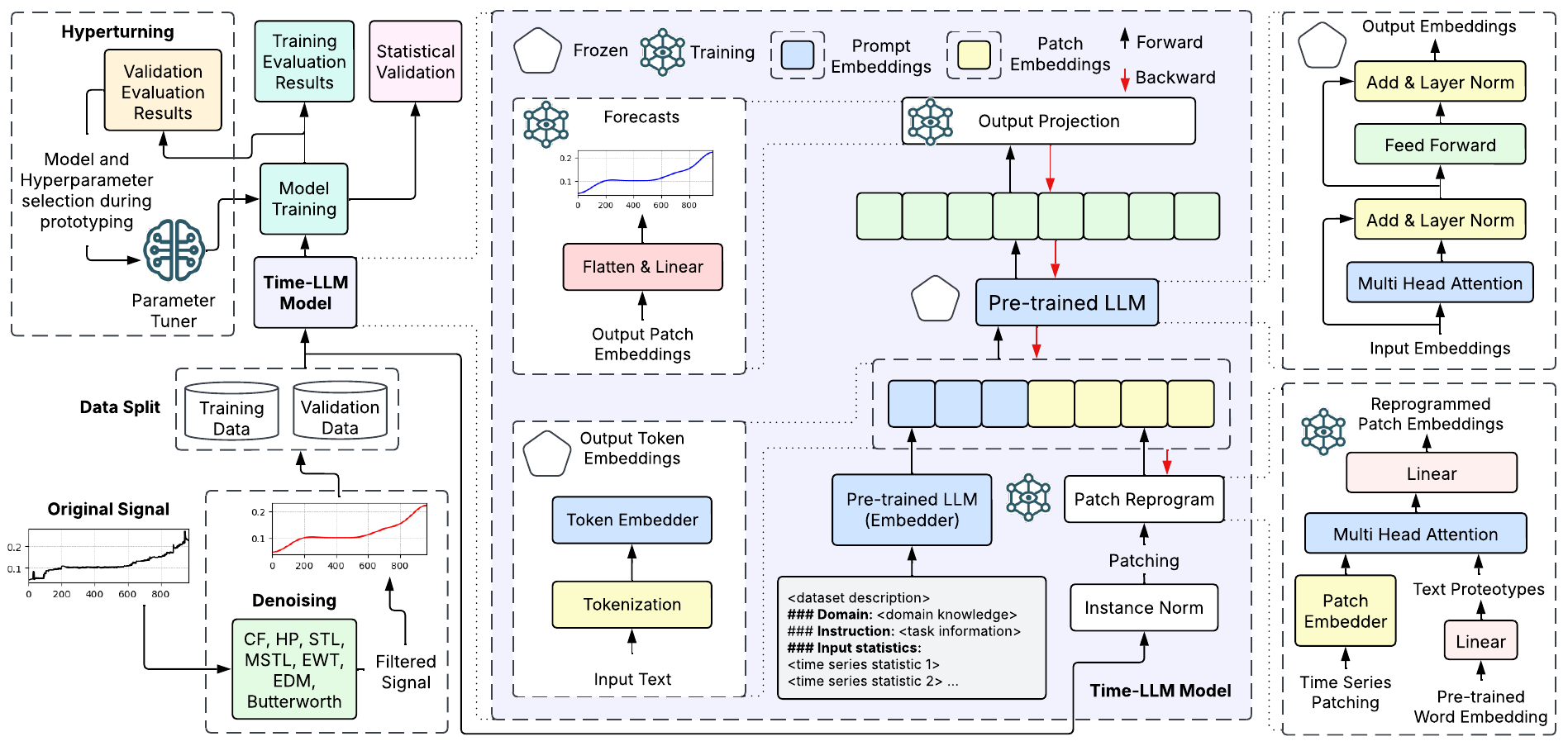}}
	\caption{\label{Fig-flow}Proposed optimized LLM model.}
\end{figure*}

In this paper, the Optuna framework is considered for the application of the TPE algorithm. The Monte Carlo approach is considered for analyzing different forecast horizons, this method relies on generating random numbers to explore a wide range of possible scenarios within a model, ensuring adequate random sampling~\citep{lye2021sampling}. The complete pipeline of the proposed optimized LLM method is presented in Figure~\ref{Fig-flow}.

\section{Results and Discussion} \label{4}

This section presents the dataset used for the experiments, the analysis setup, the results of applying the proposed optimized LLM, and a benchmarking with other well-established DL methods. After defining the dataset and comparison settings used in this paper, the results of applying filters to reduce noise are discussed. From the filtered signal, the model is optimized using hypertuning. Once optimized, an analysis is handled concerning the variation in the forecast horizon, after which comparative results of the proposed model to other models are presented.

\subsection{Dataset} \label{dataset}

The data set considered in this paper refers to leakage current measurements in an experiment on insulators subjected to artificial contamination. The experiment consists of increasing the level of contamination in insulators until a disruptive discharge occurs. The increase in contamination results in an increase in leakage current, which is the main indicator that a fault may occur \citep{IEC}. 
Six insulators were evaluated in the experiment, four of which were discharged before the end of the experiment and were disregarded. Of the two insulators that withstood the increase in contamination without being discharged, only one had a linear increase in leakage current, and this was the insulator considered in this study. 

\begin{figure*}[htb!]
	\centering
	\setkeys{Gin}{width=0.49\textwidth}{\includegraphics{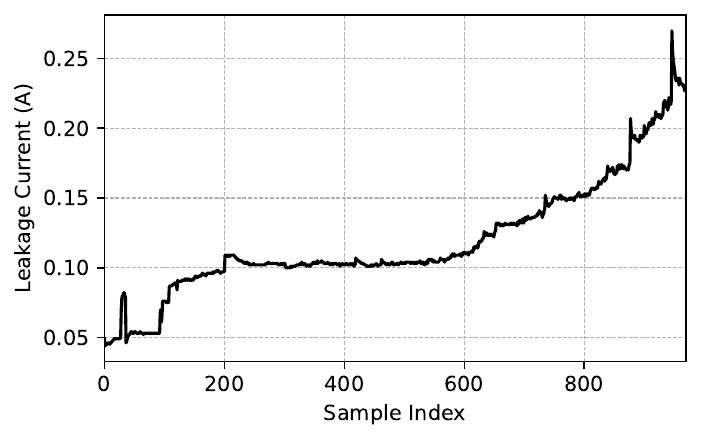}}
	\caption{\label{Fig-data}Leakage current in an insulator with artificial contamination.}
\end{figure*}

To reduce the complexity of the analysis, a downsample is handled in the pre-preprocessing stage, which means that instead of 96,800 recorded records being considered, corresponding to 26.9 hours of evaluation, 968 are the focus of the analysis. Electrical discharges occurred in many insulators after the leakage current exceeded 200mA, which is an appropriate threshold for predicting that a fault will occur. The considered signal of the leakage current in the insulator in question is shown in Figure~\ref{Fig-data}. 

The artificial contamination considered was applied in a high-voltage laboratory, in which a voltage of 8.66 kV RMS 60Hz was applied. The increase in contamination was handled following the \citet{IEC} standard, which is specific to this evaluation. LabVIEW software was used to monitor and save the leakage current values from the experiment. The insulators were fixed according to the power utility's standards and were grounded to measure the current flowing through the ground. 

\subsection{Experiment Setup} \label{Performance}

The experiments utilized an NVIDIA RTX 3060 TI graphics processing unit with 120 GB of random-access memory. The models were implemented in Python. Processing time encompasses the total duration required for both model training and testing. The evaluation metrics include RMSE, MAE, MAPE, and SMAPE, defined as follows:

\begin{equation}
 \mathrm{RMSE} = \sqrt{\frac{1}{n}\sum_{i=1}^{n}(y_i - \hat{y}_i)^2},
 \label{eq:rmse}
\end{equation}

\begin{equation}
 \mathrm {MAE}= \frac{1}{{n}} \sum _{i=1}^{n} {|y_i-\hat{y}_i|},
\end{equation}

\begin{equation}
 \mathrm {MAPE}= \frac{100\%}{n} \sum_{t=1}^{n} \left| \frac{y_i - \hat{y}_i}{y_i} \right|,
\end{equation}

\begin{equation}
 \mathrm{SMAPE} = \frac{100\%}{n} \sum_{i=1}^{n} \frac{|y_i - \hat{y}_i|}{(|{y_i} + \hat{y}_i|)/2},
\end{equation}

\noindent where $y_i$ is the actual value, $\hat{y}$ is the forecasted value, and $n$ is the number of observations.

%\subsection{Comparative Analysis}

To have a comparative assessment of the proposed optimized LLM, the standard RNN, dilated RNN, LSTM, GRU, TFT, TCN, informer, DeepNPTS, N-BEATS, and NHITS models are evaluated. In the analysis of different forecast horizons, considering multiple LLM models, the Monte Carlo approach was applied with a step equal to 5, from the initial horizon to the last horizon evaluated. 

The comparative analysis considered multi-horizon prediction, taking into account a horizon of 5 to 60 steps ahead (short-term and medium-term horizons). Each sample in the experiment considers the signal recorded at a time step of one second. The experiment, which lasted a few hours, represents the accumulation of contaminants that would occur in an external environment over approximately thirty years. The time it takes for a fault to develop depends heavily on the characteristics of the location where the system is installed.

\subsection{Filtering Analysis}

\begin{figure*}[htb!]
	\centering
	\setkeys{Gin}{width=0.49\textwidth}{\includegraphics{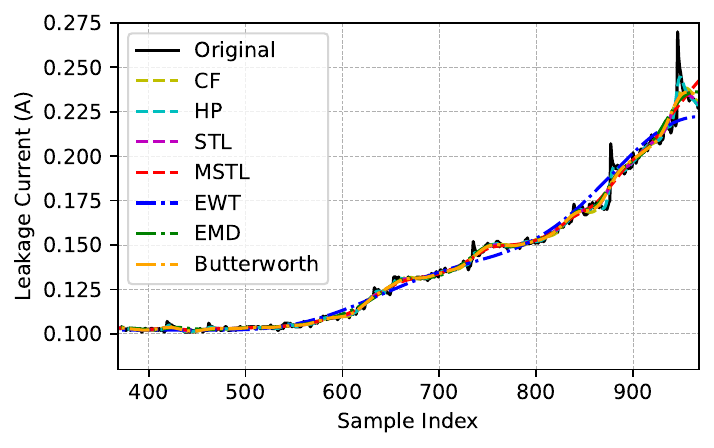}}
	\caption{\label{Fig-filters}Comparison of application of filters to the original signal.}
\end{figure*}

In the pre-processing stage, to ensure a fair analysis, all the filters considered in this paper are applied using their default settings. The results of this evaluation in relation to the original signal are shown in Figure~\ref{Fig-filters}. This figure shows the sample index as the horizontal axis, since the downsample is applied to reduce the complexity of the analysis. %, as explained in Section~\ref{dataset}.

Based on these results, the timeLLM model is applied to the original signal and the denoised signals by all the filters, as shown in Table~\ref{filver_eval}. In this analysis, the default setup is used in the model considering a horizon equal to 60 steps ahead. The time needed to process the model with each of the filters was equivalent since the greatest computational effort is regarding the training stage of the DL model, not the usage of the filter.

\begin{table}[htb!]
%\footnotesize
\small
\caption{Results of timeLLM using different filters.}
\label{filver_eval}
\begin{tabular}{llllllllll}
\hline
Filter & RMSE & MAE & MAPE & SMAPE & Time (s) \\ 
\hline
Original & 6.06$\times10^{-3}$ & 4.63$\times10^{-3}$ & 3.33$\times10^{-2}$ & 3.42 & 1.24$\times10^{1}$\\
CF	& 5.52$\times10^{-3}$ & 4.32$\times10^{-3}$ & 3.12$\times10^{-2}$ & 3.20 & 1.27$\times10^{1}$\\
HP	& 5.38$\times10^{-3}$ & 4.23$\times10^{-3}$ & 3.05$\times10^{-2}$ & 3.13 & 1.28$\times10^{1}$\\
STL	& 5.15$\times10^{-3}$ & 3.95$\times10^{-3}$ & 2.85$\times10^{-2}$ & 2.92 & \textbf{1.23$\times10^{1}$}\\
MSTL & 5.63$\times10^{-3}$ & 4.56$\times10^{-3}$ & 3.29$\times10^{-2}$ & 3.38 & 1.25$\times10^{1}$\\
EWT	& \textbf{3.15$\times10^{-3}$} & \textbf{2.99$\times10^{-3}$} & \textbf{2.25$\times10^{-2}$} & \textbf{2.28} & 1.25 $\times10^{1}$\\
EMD	& 5.27$\times10^{-3}$ & 3.69$\times10^{-3}$ & 2.64$\times10^{-2}$ & 2.71 & \textbf{1.23$\times10^{1}$}\\
Butterworth	& 5.04$\times10^{-3}$ & 3.86$\times10^{-3}$ & 2.79$\times10^{-2}$ & 2.86 & 1.28$\times10^{1}$\\
\hline
\multicolumn{3}{l}{Best results in bold}\\
\end{tabular} \centering
\end{table}

Considering the pre-defined setup of the filtering techniques, the best results were found using EWT, as the RMSE is equal to 3.15$\times10^{-3}$, MAE equal to 2.99$\times10^{-3}$, MAPE equal to 2.25\%, and a SMAPE equal to 2.28. It is important to note that all these filters could be applied at this stage, but fine-tuning is required for each specific filter. 

The results of using all the filters were superior to the results obtained by the base model without using the filter, showing that their application is promising where the model using the EWT filter showed, in general, lower errors than the other filters and the original data. In the following subsection, the model hypertuning evaluation is presented.

\subsection{Hypertuning Analysis}

An important definition to be made in the model's tuning phase is the variability space (gap) of each hyperparameter. If the space is not properly defined, the model may struggle to find the optimum. A large search space makes it difficult to optimize the hyperparameters, and a small space can limit the search, causing the TPE to find values near the extreme of variation for each hyperparameter. 

\begin{figure*}[htb!]
	\centering
	\setkeys{Gin}{width=1\textwidth}{\includegraphics{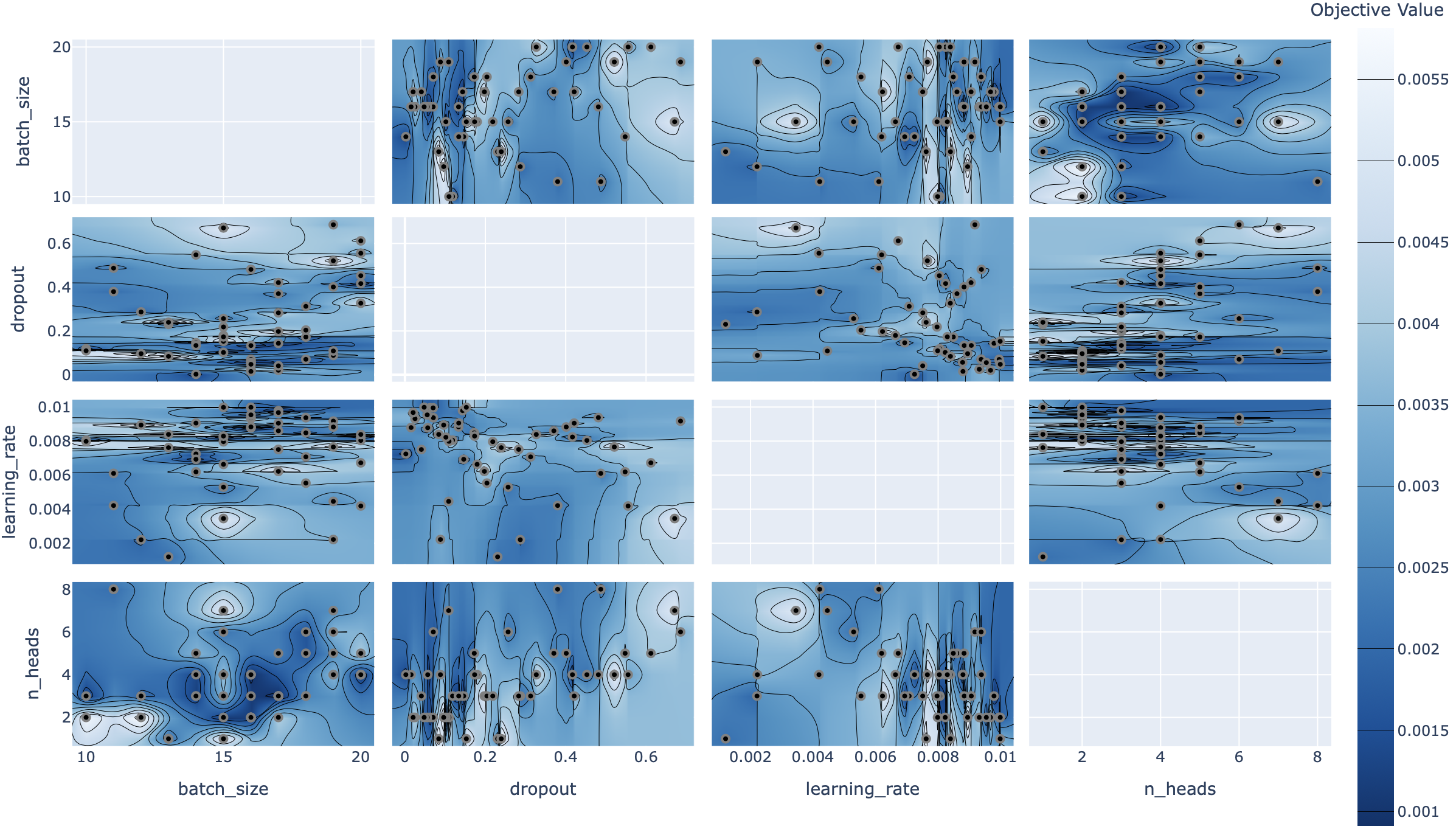}}
	\caption{\label{Fig-contour}Results of multi-objective optimization.}
\end{figure*}

To ensure that the model has been properly optimized, hypertuning with a large search space has been done beforehand, so the search space presented here takes into account the gap of best values found in the first experiment.
Considering the gap of the first experiment, the hypertuning is done with the following values of batch size [10 to 20], number of heads [1 to 8], learning rate [0.001 to 0.01], and dropout [0.0 to 0.7]. The results of this analysis are shown in Figure~\ref{Fig-contour}. 
This analysis considered a time-LLM model with the EWT filter, using a horizon and input size equal to 20. Considering that all the optimal results of the hyperparameters evaluated were within the variation gap analyzed, the analysis was carried out properly.

In Figure~\ref{Fig-contour}, the black dots represent the local minima of the gradient of each combination, considering Eq.~(\ref{eq:rmse}) as the loss function. 
The lighter the gradient (white being the lightest color), the greater the loss function result and, in turn, the worse the result obtained. Conversely, the darker the color of the gradient (in this case dark blue), the lower the loss function result and, in turn, the better the result, since in both cases the optimization function is to reduce the RMSE.

In this optimization, the batch size hyperparameter had an importance of 13\% in achieving the goal of reducing the RMSE, the learning rate had an importance of 17\%, the number of heads in the model had an importance of 19\%, and the dropout had an importance of 51\%, making it the most important hyperparameter to optimize. 

\begin{figure}[htb!]
	\centering
	\setkeys{Gin}{width=1\textwidth}{\includegraphics{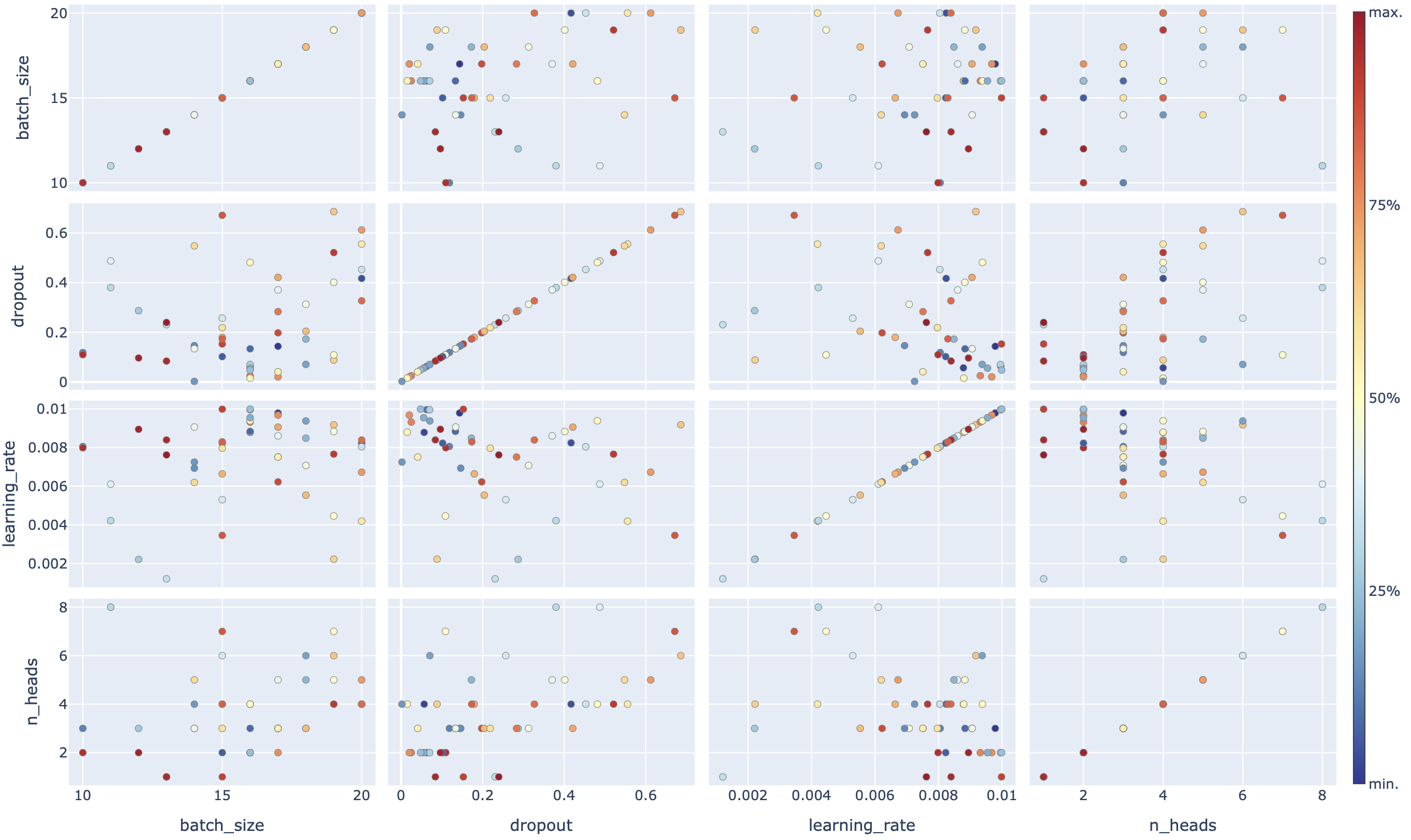}}
	\caption{\label{Fig-rank}Trial rank of hyperparâmeters.}
\end{figure}

The rank of the hyperparameter fitting attempts is shown in Figure~\ref{Fig-rank}. The values in the center of the plots are linear, as they only show the variation of one hyperparameter at a time. It should be noted that, for this case, the redder it gets, the higher the loss function (RMSE in this case), and, in turn, the worse the approximation of the output to the ground truth. Similarly, the lower the value of the loss function, the color will tend toward dark blue, and the closer the output model will be to the ground truth.

Multi-criteria optimization tries interactively to find the best hyperparameter setting to minimize the objective function, reducing the RMSE in this paper, by finding an optimum that combines the setting of several hyperparameters simultaneously. %These trials are shown in Figure~\ref{Fig-parallel}. 
The optimum values found in this experiment were a batch size of 17, using 3 heads, a learning rate of 9.77$\times10^{-3}$, and a dropout equal to 1.43$\times10^{-1}$, this hyperparameter setup was used for the following analyses, this being the optimized model.

\begin{figure}[htb!]
	\centering
	\setkeys{Gin}{width=0.49\textwidth}{\includegraphics{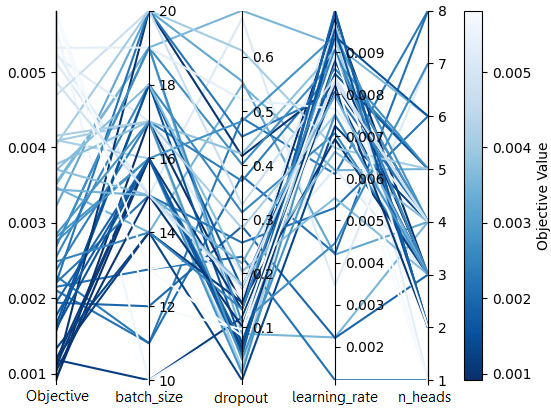}}
	\caption{\label{Fig-parallel}Model trails in hypertuning.}
\end{figure}

Figure~\ref{Fig-parallel} presents the different combinations between different variables and the lighter the color of the combination, the higher the loss function value and, consequently, the worse the model result. The darker the color of the combination between different variables (dark blue in this case), the lower the loss function value and therefore the better the model output.

An interesting observation in this presentation (Figure~\ref{Fig-parallel}) is that many values are close in the trials. This shows an advantage in the application of TPE since when the value of the hyperparameter is getting close to the optimum, the algorithm creates new, more assertive trials for multi-criteria optimization. Other hypertuning techniques, such as random sampling, can have a larger search space, but because they are random trials, fewer values close to the optimum are evaluated.

\subsection{Multi-horizon Analysis}

The purpose of the evaluation of horizons is to analyze the impact using a longer forecast horizon has on the model's performance. To this end, Table~\ref{horizon_eval} presents a comparative analysis of various forecast horizons using the proposed optimized LLM model using the EWT as the input stage filter. A forecast of 5 steps ahead is considered a short-term horizon, and 60 steps ahead is considered a medium-term horizon.

\begin{table}[htb!]
%\footnotesize
\small
\caption{Time step horizon analysis.}
\label{horizon_eval}
\begin{tabular}{llllllllll}
\hline
Horizon & RMSE & MAE & MAPE & SMAPE & Time (s) \\ 
\hline
05 & \textbf{2.24$\times10^{-4}$} & \textbf{1.84$\times10^{-4}$} & \textbf{1.41$\times10^{-3}$} & \textbf{1.41$\times10^{-1}$} & \textbf{1.67$\times10^{1}$}\\
10 & 3.58$\times10^{-4}$ & 2.95$\times10^{-4}$ & 2.25$\times10^{-3}$ & 2.25$\times10^{-1}$ & 1.68$\times10^{1}$\\
15 & 4.50$\times10^{-4}$ & 3.84$\times10^{-4}$ & 2.90$\times10^{-3}$ & 2.91$\times10^{-1}$ & \textbf{1.67$\times10^{1}$}\\
20 & 4.62$\times10^{-4}$ & 3.77$\times10^{-4}$ & 2.84$\times10^{-3}$ & 2.85$\times10^{-1}$ & 1.68$\times10^{1}$\\
25 & 5.79$\times10^{-4}$ & 4.98$\times10^{-4}$ & 3.76$\times10^{-3}$ & 3.77$\times10^{-1}$ & 1.71$\times10^{1}$\\
30 & 2.82$\times10^{-3}$ & 2.61$\times10^{-3}$ & 1.95$\times10^{-2}$ & 1.97$\times10^{-1}$ & \textbf{1.67$\times10^{1}$}\\
35 & 5.87$\times10^{-4}$ & 4.40$\times10^{-4}$ & 3.27$\times10^{-3}$ & 3.26$\times10^{-1}$ & 1.70$\times10^{1}$\\
40 & 1.32$\times10^{-3}$ & 1.07$\times10^{-3}$ & 7.86$\times10^{-3}$ & 7.82$\times10^{-1}$ & 1.68$\times10^{1}$\\
45 & 1.41$\times10^{-3}$ & 1.21$\times10^{-3}$ & 8.90$\times10^{-3}$ & 8.85$\times10^{-1}$ & \textbf{1.67$\times10^{1}$}\\
50 & 1.02$\times10^{-3}$ & 8.54$\times10^{-4}$ & 6.27$\times10^{-3}$ & 6.25$\times10^{-1}$ & 1.68$\times10^{1}$\\
55 & 1.54$\times10^{-3}$ & 1.19$\times10^{-3}$ & 8.66$\times10^{-3}$ & 8.60$\times10^{-1}$ & 1.70$\times10^{1}$\\
60 & 1.21$\times10^{-3}$ & 9.30$\times10^{-4}$ & 6.77$\times10^{-3}$ & 6.73$\times10^{-1}$ & 1.68$\times10^{1}$\\
\hline
\multicolumn{3}{l}{Best results in bold}\\
\end{tabular} \centering
\end{table}

The results of this comparison showed that the longer the forecast horizon, the more difficult it is to produce a forecast with lower error. The time needed to train the model and carry out the test did not vary considerably when changing the forecast horizon, but the error had a big impact, showing how challenging it is to carry out forecasts with long forecast horizons.

The MAPE values show that the model struggles to make predictions for horizons greater than 20 steps ahead since the values were higher than 30\%. This challenge is related to the variability of the data, where non-linear time series are more difficult to predict in horizons that consider many steps ahead. This result shows that the proposed model is better suited to short-term than medium-term horizons.

\begin{figure*}[htb!]
	\centering
	\setkeys{Gin}{width=0.49\textwidth}{\includegraphics{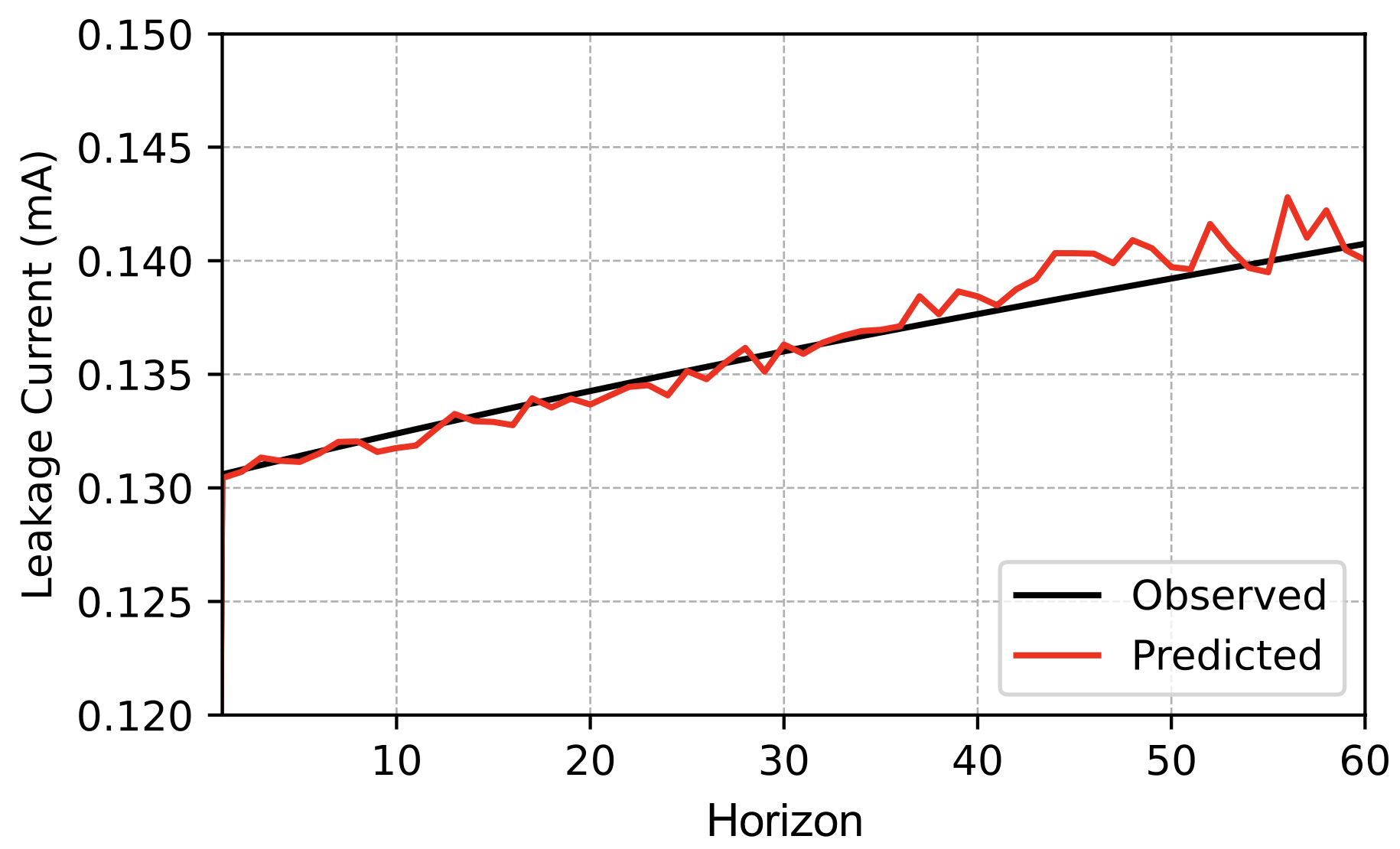}}
	\caption{\label{Fig-hor}Forecast results versus the original signal for different prediction horizons.}
\end{figure*}

A visualization of the variation in the model's performance as the forecast horizon increases can be seen in Figure~\ref{Fig-hor}. 
For random sampling, a Monte Carlo approach is considered to train the model. The model shows a considerably promising result up to a Horizon equal to 20 steps ahead, after which greater difficulties are encountered in handling the forecast, especially after 35 steps ahead, this result is expected because the longer the horizon, the more difficult it becomes to predict the variation.

\subsection{Statistical Analysis}

To perform a statistical analysis, the proposed optimized LLM model was computed for 50 runs with different initialization weights (seed). The result of this evaluation is shown in Table~\ref{table_sta}. The mean, median, mode, range, standard deviation, 25th, 50th, and 75th percentile (\%ile), interquartile range (IQR), skewness, and kurtosis for each performance metric (RMSE, MAE, MAPE and SMAPE) are presented.
In general, the results of the statistical analysis show that the model has promising average values, making it a suitable model for the task discussed here.

\begin{table}[htb!]
%\footnotesize
\small
\caption{Statistical performance of the hypertuned
EWT-LLM considering 50 runs.}
\label{table_sta}
\begin{tabular}{llllll}
\hline
{\color{black}Metrics} & RMSE & MAE & MAPE & SMAPE \\
\hline
Mean & 1.61$\times10^{-3}$ & 1.38$\times10^{-3}$ & 1.00$\times10^{-2}$ & 1.01 \\
Median & 1.06$\times10^{-3}$ & 8.50$\times10^{-4}$ & 6.30$\times10^{-3}$ & 6.32$\times10^{-1}$ \\
Mode & 4.40$\times10^{-4}$ & 3.60$\times10^{-4}$ & 2.66$\times10^{-3}$ & 2.65$\times10^{-1}$ \\
Range & 5.25$\times10^{-3}$ & 4.63$\times10^{-3}$ & 3.37$\times10^{-2}$ & 3.46 \\
Std. Dev. & 1.40$\times10^{-3}$ & 1.25$\times10^{-3}$ & 9.07$\times10^{-3}$ & 9.24$\times10^{-1}$ \\
25th \%ile & 6.70$\times10^{-4}$ & 5.30$\times10^{-4}$ & 3.91$\times10^{-3}$ & 3.91$\times10^{-1}$ \\
50th \%ile & 1.06$\times10^{-3}$ & 8.50$\times10^{-4}$ & 6.30$\times10^{-3}$ & 6.32$\times10^{-1}$ \\
75th \%ile & 1.82$\times10^{-3}$ & 1.56$\times10^{-3}$ & 1.13$\times10^{-2}$ & 1.14 \\
IQR & 1.15$\times10^{-3}$ & 1.02$\times10^{-3}$ & 7.42$\times10^{-3}$ & 7.46$\times10^{-1}$ \\
Skewness & 1.65 & 1.66 & 1.66 & 1.70 \\
Kurtosis & 1.81 & 1.88 & 1.89 & 2.05 \\
\hline
\end{tabular} \centering
\end{table}

\begin{figure*}[htb!]
	\centering
	\setkeys{Gin}{width=0.49\textwidth}{\includegraphics{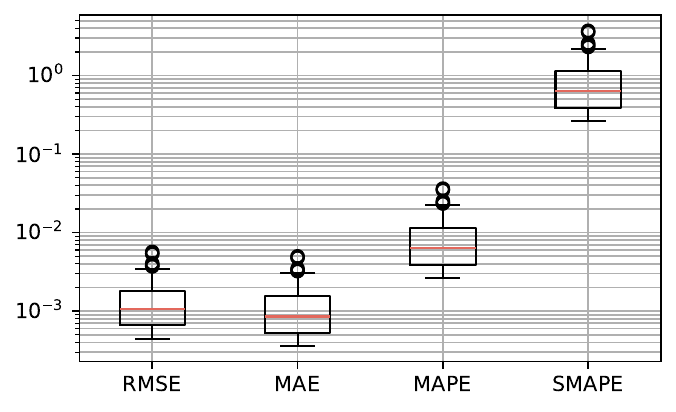}}
	\caption{\label{Fig-violinplots}Results of RMSE, MAE, MAPE, and SMAPE considering 50 runs.}
\end{figure*}

To give a visual presentation of the variability of the results in relation to the 50 runs, Figure~\ref{Fig-violinplots} shows a box plot of the errors evaluated in this paper using the proposed optimized LLM model. This presentation shows some outliers with greater error than the average values obtained by the model. 

%\begin{figure*}[htb!]
%	\centering
%	\setkeys{Gin}{width=1.0\textwidth}{\includegraphics{Figures/violinplots.pdf}}
%	\caption{\label{Fig-violinplots}Violin results according to RMSE, MAE, MAPE, and SMAPE.}
%\end{figure*}

Considering that the initialization of the weights is random and that the vast majority of the results are close to the average (observing a logarithmic presentation), the results are promising. These results show that the model is stable, and based on that, a final comparative analysis is presented in the following.

\subsection{Benchmarking}

The comparative analysis presented here focuses on comparing our proposed method to other well-established DL architectures. This evaluation considers two different horizons: A short-term horizon of 5-steps ahead and a medium-term horizon of 60-steps ahead.
The results of the short-term horizon are presented in Table~\ref{ben1}, and the results of the medium-term horizon are presented are presented in Table~\ref{ben2}.

\begin{table}[htb!]
%\footnotesize
\small
\caption{Comparative analysis of DL models for a short-term horizon.}
\label{ben2}
\begin{tabular}{llllllllll}
\hline
Model & RMSE & MAE & MAPE & SMAPE & Time (s) \\ 
\hline
Standard RNN & 5.41$\times10^{-3}$ & 5.09$\times10^{-3}$ & 3.87$\times10^{-2}$ & 3.96 & \textbf{3.27} \\
Dilated RNN & 3.48$\times10^{-3}$ & 3.11$\times10^{-3}$ & 2.37$\times10^{-2}$ & 2.40 & 9.69 \\
LSTM & 2.06$\times10^{-3}$ & 1.95$\times10^{-3}$ & 1.48$\times10^{-2}$ & 1.49 & 3.65 \\
GRU & 1.87$\times10^{-3}$ & 1.71$\times10^{-3}$ & 1.30$\times10^{-2}$ & 1.29 & 4.12 \\
TFT & 7.20$\times10^{-4}$ & 5.81$\times10^{-4}$ & 4.43$\times10^{-3}$ & 4.42$\times10^{-1}$ & 1.08$\times10^{2}$ \\
TCN & 5.87$\times10^{-3}$ & 4.85$\times10^{-3}$ & 3.69$\times10^{-2}$ & 3.59 & 4.03 \\
Informer & 1.22$\times10^{-2}$ & 9.33$\times10^{-3}$ & 7.09$\times10^{-2}$ & 7.30 & 7.88$\times10^{1}$ \\
DeepNPTS & 6.67$\times10^{-4}$ & 6.02$\times10^{-4}$ & 4.59$\times10^{-3}$ & 4.58$\times10^{-1}$ & 3.34 \\
N-BEATS & 8.97$\times10^{-4}$ & 7.48$\times10^{-4}$ & 5.71$\times10^{-3}$ & 5.69$\times10^{-1}$ & 3.36$\times10^{1}$ \\
NHITS & 8.74$\times10^{-4}$ & 7.34$\times10^{-4}$ & 5.60$\times10^{-3}$ & 5.58$\times10^{-1}$ & 4.89$\times10^{1}$ \\
\hline
Our & \textbf{2.24$\times10^{-4}$} & \textbf{1.84$\times10^{-4}$} & \textbf{1.41$\times10^{-3}$} & \textbf{1.41$\times10^{-1}$} & 1.67$\times10^{1}$\\
\hline
\multicolumn{3}{l}{Best results in bold}\\
\end{tabular} \centering
\end{table}

For a short-term horizon, the most promising results were considering the TFT and the DeepNPTS that had an RMSE of 7.20$\times10^{-4}$ and 6.67$\times10^{-4}$ respectively. In this evaluation, the TFT needed a longer time to be computed, which is not an issue here, considering that the analysis is performed offline, and the time required for training is just for computation comparison. Our proposed model (optimized LLM) outperforms all these state-of-the-art models, having an RMSE of 2.24$\times10^{-4}$ for a short-term horizon.

\begin{table}[htb!]
%\footnotesize
\small
\caption{Comparative analysis of DL models for a medium-term horizon.}
\label{ben1}
\begin{tabular}{llllllllll}
\hline
Model & RMSE & MAE & MAPE & SMAPE & Time (s) \\ 
\hline
Standard RNN & 1.42$\times10^{-2}$ & 1.34$\times10^{-2}$ & 9.77$\times10^{-2}$ & 1.03$\times10^{1}$ & 6.07 \\
Dilated RNN & 2.69$\times10^{-2}$ & 2.64$\times10^{-2}$ & 1.94$\times10^{-1}$ & 2.15$\times10^{1}$ & 1.40$\times10^{1}$ \\
LSTM & 2.19$\times10^{-2}$ & 2.11$\times10^{-2}$ & 1.54$\times10^{-1}$ & 1.68$\times10^{1}$ & \textbf{5.81} \\
GRU & 1.21$\times10^{-2}$ & 9.32$\times10^{-3}$ & 6.71$\times10^{-2}$ & 7.10 & 1.15$\times10^{1}$ \\
TFT & 1.97$\times10^{-2}$ & 1.68$\times10^{-2}$ & 1.23$\times10^{-1}$ & 1.13$\times10^{1}$ & 1.31$\times10^{3}$ \\
TCN & 3.59$\times10^{-2}$ & 3.41$\times10^{-2}$ & 2.53$\times10^{-1}$ & 2.22$\times10^{1}$ & 5.96 \\
Informer & 1.05$\times10^{-1}$ & 5.10$\times10^{-2}$ & 3.73$\times10^{-1}$ & 3.79$\times10^{1}$ & 8.02$\times10^{2}$ \\
DeepNPTS & 7.40$\times10^{-3}$ & 6.01$\times10^{-3}$ & 4.34$\times10^{-2}$ & 4.48 & 8.81 \\
N-BEATS & 5.22$\times10^{-3}$ & 3.73$\times10^{-3}$ & 2.68$\times10^{-2}$ & 2.73 & 4.00$\times10^{1}$ \\
NHITS & 6.94$\times10^{-3}$ & 4.95$\times10^{-3}$ & 3.55$\times10^{-2}$ & 3.67 & 5.11$\times10^{1}$ \\
\hline
Our & \textbf{1.21$\times10^{-3}$} & \textbf{9.30$\times10^{-4}$} & \textbf{6.77$\times10^{-3}$} & \textbf{6.73$\times10^{-1}$} & 1.68$\times10^{1}$\\
\hline
\multicolumn{3}{l}{Best results in bold}\\
\end{tabular} \centering
\end{table}

For a medium-term horizon (horizon = 60), as shown in Table~\ref{ben1}, the state-of-the-art models that showed the best results were DeepNPTS, N-BEATS, and NHITS with an RMSE below than $\times10^{-2}$. The algorithm proposed in this paper had better results with an RMSE of $1.21\times10^{-3}$, which clearly shows better performance in predicting future samples.

\section{Conclusion} \label{5}

In this paper, a hybrid model integrating a filtering input stage and an LLM applied for time series, optimized via the Optuna framework, was proposed for insulator fault prediction. The time series of the leakage current of the insulators is based on a high-voltage experiment with artificial contamination.
Considering an evaluation of CF, HP, STL, MSTL, EWT, Butterworth, and EDM filters, the EWT showed more promising results with its default setup. Therefore, the EWT decomposed time series signals, mitigating noise and non-stationary effects. These decomposed signals were subsequently forecasted by an optimized LLM, which demonstrated robust capabilities in modeling complex temporal dependencies inherent in insulator degradation patterns. 

The experimental results underscore the superiority of the optimized LLM framework in comparison to the state-of-the-art DL models, achieving an RMSE equal to 2.24$\times10^{-4}$ for a short-term horizon (5 steps ahead) and 1.21$\times10^{-3}$ for a medium-term horizon (60 steps ahead) in insulator fault detection. The proposed methodology enhances the reliability of power grid maintenance and provides a generalizable framework for predictive maintenance in industrial systems.

Future work could explore the integration of real-time adaptive decomposition techniques, expand the model’s interpretability for operational decision-making, and validate the framework on larger cross-domain datasets. Deploying the monitoring model on-site can promote practical implementation in smart grid applications.

\section*{Acknowledgements}
\noindent 
This work was partially supported by the Natural Sciences and Engineering Research Council of Canada (NSERC), funding reference number DDG-2024-00035. Cette recherche a été financée par le Conseil de recherches en sciences naturelles et en génie du Canada (CRSNG), numéro de référence DDG-2024-00035. 
This research was also partially funded by the Funda\c c\~ao para a Ci\^encia e a Tecnologia (FCT, \url{https://ror.org/00snfqn58}) under Grants UIDB/04111/2020, UIDB/00066/2020, UIDB/00408/2020 and the LASIGE Research Unit, ref. UID/000408/2025, as well as by the Instituto Lusófono de Investigação e Desenvolvimento (ILIND), Portugal, under Project COFAC/ILIND\-/\-COPELABS\-/1/2024.
A realiza\c{c}\~ao desta investiga\c{c}\~ao foi também parcialmente financiada por fundos nacionais atrav\'es da FCT - Funda\c{c}\~ao para a Ci\^encia e Tecnologia, I.P. no \^ambito dos projetos UIDB/04466/2020, UIDP/04466/2020. (in Portuguese)

\section*{Declaration of competing interest}
The authors declare that they have no known competing financial interests or personal relationships that could have appeared to influence the work reported in this paper.

\section*{CRediT authorship contribution statement}

\textbf{João Pedro Matos-Carvalho} and \textbf{Stefano Frizzo Stefenon}: Conceptualization, software, formal analysis, writing - original draft. \textbf{Valderi Reis Quietinho Leithard} and \textbf{Kin-Choong Yow}: Supervision, writing - review \& editing.

\section*{Data availability}
% Data will be made available on request.
For future analysis, the original data is available at: \url{https://github.com/SFStefenon/LeakageCurrent}.

\bibliographystyle{unsrtnat}
%\bibliography{biblio}

%% else use the following coding to input the bibitems directly in the
%% TeX file.

%\begin{thebibliography}{00}

%% \bibitem[Author(year)]{label}
%% Text of bibliographic item

%\bibitem[ ()]{}

%\end{thebibliography}
\end{document}